\documentclass{article}

\usepackage{arxiv}

\usepackage[utf8]{inputenc} 
\usepackage[T1]{fontenc}    
\usepackage{hyperref}       
\usepackage{url}            
\usepackage{booktabs}       
\usepackage{amsfonts}       
\usepackage{nicefrac}       
\usepackage{microtype}      
\usepackage{lipsum}		
\usepackage{graphicx}
\usepackage{natbib}
\usepackage{doi}
\usepackage{placeins}
\usepackage{palatino,epsfig,latexsym}
\usepackage[acronym]{glossaries}
\usepackage{bm}

\makeglossaries
\usepackage{algorithm,algorithmicx,algpseudocode}
\usepackage[acronym]{glossaries}
\newacronym{EA}{EA}{evolutionary algorithm}
\newacronym{QD}{QD}{quality diversity}
\newacronym{CFD}{CFD}{computational fluid dynamics}
\newacronym{VAE}{VAE}{variational autoencoder}
\newacronym{CA}{CA}{cellular automata}
\newacronym{PD}{PD}{pure diversity}
\newacronym{CPPN}{CPPN}{compositional pattern producing network}
\newacronym{NEAT}{NEAT}{neuroevolution of augmented topologies}
\newacronym{MAP-Elites}{MAP-Elites}{multidimensional archive of phenotypic elites}
\newacronym{t-SNE}{t-SNE}{t-distributed stochastic neighborhood embedding}
\newacronym{CAD}{CAD}{computer-aided design}
\title{On the Suitability of Representations for Quality Diversity Optimization of Shapes}

\date{April 6, 2023}	

\author{ 
\href{https://orcid.org/0000-0003-0174-2543}{\includegraphics[scale=0.06]{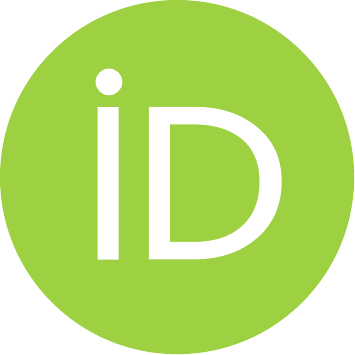}\hspace{1mm}Ludovico~Scarton}\thanks{This is the final version and has been accepted for publication at the GECCO conference} \\
Department of Computer Science \\
Bonn-Rhein-Sieg University of Applied Sciences \\
Sankt Augustin, 53757, Germany \\
	\texttt{ludovico.scarton@smail.inf.h-brs.de} \\
	\And
\href{https://orcid.org/0000-0002-8668-1796}{\includegraphics[scale=0.06]{orcid.pdf}\hspace{1mm}Alexander~Hagg}\\
Institute of Technology, Resource and Energy-efficient Engineering (TREE) \\
Bonn-Rhein-Sieg University of Applied Sciences \\
Sankt Augustin, 53757, Germany \\
\texttt{alexander.hagg@h-brs.de} \\
}

\renewcommand{\shorttitle}{On the Suitability of Representations for Quality Diversity Optimization of Shapes}

\hypersetup{
pdftitle={On the Suitability of Representations for Quality Diversity Optimization of Shapes},
pdfsubject={cs.NE},
pdfauthor={Ludovico Scarton, Alexander Hagg},
pdfkeywords={encoding, representation, quality diversity, compositional pattern producing networks, cellular automata, parametric},
}

\begin{document}
\maketitle

\begin{abstract}
	The representation, or encoding, utilized in evolutionary algorithms has a substantial effect on their performance. Examination of the suitability of widely used representations for quality diversity optimization (QD) in robotic domains has yielded inconsistent results regarding the most appropriate encoding method. Given the domain-dependent nature of QD, additional evidence from other domains is necessary. This study compares the impact of several representations, including direct encoding, a dictionary-based representation, parametric encoding, compositional pattern producing networks, and cellular automata, on the generation of voxelized meshes in an architecture setting. The results reveal that some indirect encodings outperform direct encodings and can generate more diverse solution sets, especially when considering full phenotypic diversity. The paper introduces a multi-encoding QD approach that incorporates all evaluated representations in the same archive. Species of encodings compete on the basis of phenotypic features, leading to an approach that demonstrates similar performance to the best single-encoding QD approach. This is noteworthy, as it does not always require the contribution of the best-performing single encoding.
	
\end{abstract}




\maketitle

\section{Introduction}
\label{sec:introduction}

\begin{figure}
	\includegraphics[width=1\linewidth]{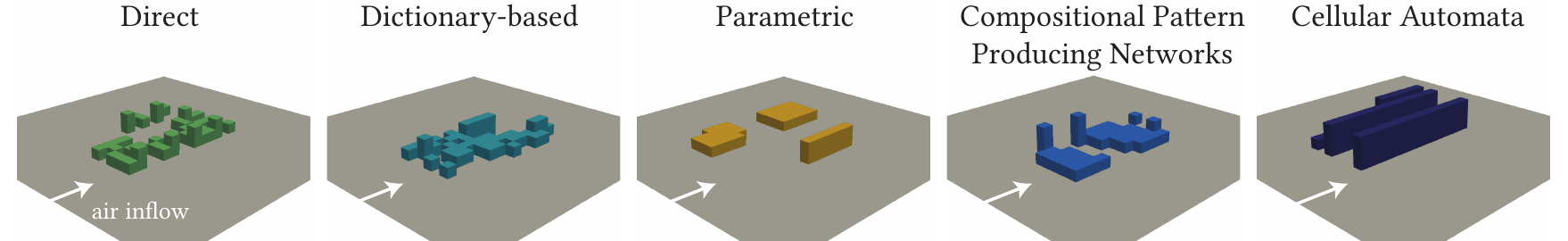}
	\caption{Architectural massing design examples generated by \gls{QD} using five representations.}
	\label{fig:teaser}
\end{figure}

\Gls{QD} optimization is an evolutionary paradigm based on the separation of the search space, the genome, and the niching space, consisting of features that describe particular aspects of the solution, the phenotype.
The algorithms in this paradigm create diverse, high-performing solution sets. 
In doing so, \gls{QD} can provide insights into the underlying structure of the problem, offer many design variants to the user, and increase the robustness of a solution in dynamic and uncertain environments.
It is crucial that the solution encoding, also called representation, allows the creation of solution sets that are as high-quality and diverse as possible.
\gls{QD} optimization usually targets engineering solutions of high-dimensional nature, because they encompass extended phenotypes~\citep{hagg2021phenotypic} like time-dependent robot walking gait strategies~\citep{Cully2015} or the shape of the air flow around buildings~\citep{hagg2020designing}. 
Past research has shown that the encoding of solutions has a major impact on the performance of \glspl{EA} but usually focuses on the time to convergence of \glspl{EA} on one hand, and the quality of solutions on the other, as was done in the time-quality framework presented in~\citep{rothlauf2006representations}.
One important find was that the convergence time of \glspl{EA} tends to grow super-linearly with dimensionality of the encoding~\citep{chen2015measuring}.
But in contrast to other \glspl{EA}, \gls{QD} algorithms put a strong focus on the diversity of solutions.
This article aims to provide insights into the impact different encodings have not only on the quality, but also on the diversity of the solution sets generated with \gls{QD}.

A comparison between various encodings that produce neural controllers for robots~\citep{tarapore2016different} gave evidence that generative encodings might actually limit \gls{QD}, and that in their case, direct encodings more easily filled the archive of solutions and produced a solution set that was generally of higher fitness than when producing it through indirect encoding.
They provided evidence to the conclusion that locality, when small mutations produce small changes in the phenotype, which is the case with direct encodings, might be more important for \gls{QD} than for more classical \glspl{EA}, 
However, when evolving robot arm shapes, other work found that indirect encodings allow for further exploration of the design space and improved fitness~\citep{collins2019comparing}.
It is still an open question how encodings behave in \gls{QD} optimization problems like shape domains.
Does \gls{QD} benefit more from smooth mutation with direct encodings, or from faster exploration with indirect encodings?

A typical shape domain is architectural design, where architects push for freedom of creativity within the boundaries of optimal design.
During the massing design phase in architecture, many design requirements have not yet been fully defined and are discovered~\citep{maher2000model}.
While this phase allows the largest design freedom, the decisions made here constrain the creative freedom and maximum design quality in later design phases.
The massing design phase is therefore an excellent problem to be solved with \gls{QD}, to inform urban planners and architects early on in the design process.
An important quality metric of a construction project is its climate impact, for example cold air flow throughput, heat absorption during the day and heat radiation at night, or wind nuisance.
Classically, such climate impacts are considered in later stages of a design project, especially through flow analyses of a concrete design.
This is a place where \gls{QD} can be used: early on in the design process.

Due to the high dimensionality of building designs, it can be useful when the encoding maps low-dimensional genomes onto these high-dimensional phenotypes.
By restricting the dimensionality of the genome, indirect encodings can reduce the convergence time of the search process, so it is expected that indirect encodings allow finding high-quality solutions faster.
Simultaneously this means that not all phenotypes might be reached, which might lead to those encodings to produce lower diversity or even lower fitness.
A good encoding uses a small number of genetic dimensions to reach many high-performing solutions.
This alignment between genetic space and the set of useful phenotypes is domain-dependent, which provides the necessity of this article.
This article focuses on encodings that are able to generate voxelized three-dimensional meshes, a common representation of massing design in architecture. 
The impacts of a range of encodings is analyzed with respect to the resulting set diversity and quality. 
Evidence is given to help answering the following research questions:
\begin{enumerate}
	\item Do direct encodings provide higher QD performance than indirect encodings, as was concluded by \citet{tarapore2016different} \citep{tarapore2016different}? 
	\item Does low genetic dimensionality result in higher QD performance, as is suggested by \citet{chen2015measuring} \citep{chen2015measuring}?
	\item What encoding is most suitable for the domain at hand?
	\item Do different encodings cover distinct regions in phenotypic space?
	\item Does a combination of encodings in QD improve the diversity of the solution set?
\end{enumerate}

To the best of our knowledge, this is the first study to systematically compare the set diversity and other characteristics of widely used encodings for voxelized three-dimensional mesh representations for \gls{QD}. 
We believe that our findings will be of interest to research on \gls{QD} in general, and practitioners working in the area of the built environment, or shape design, specifically.
This work represents an initial study for a particular domain.

\section{Methods}
\label{sec:methods}

Let us define the terminology around \textit{representations} commonly used in \glspl{EA}, which are inspired by biological evolution.
Representations in \gls{EA} are described by the genome-phenotype decoding and phenotype-fitness mapping~\citep{rothlauf2006representations}.
A \textit{phenotype} $\textbf{p} \in \mathcal{P}$ is the expressed solution for a specific problem domain.
$\mathcal{X}$ is defined as the genetic search space of the \gls{EA}.
For continuous optimization problems, the genome $\textbf{x} \in \mathcal{X}$ usually consists of a tuple of real-valued numbers.
A phenotype $p$ is encoded into a genome $\textbf{x}$ using an encoding $\mathcal{E}: \mathcal{P} \rightarrow \mathcal{X}$.
The term \textit{encoding} can also interchangeably be used instead of representation.
To retrieve $p$ from a genome, a decoding process $\mathcal{D}: \mathcal{X} \rightarrow \mathcal{P}$ is used.

Depending on what line of thought is followed, the full phenotype in building mass design might either be the morphology of the design or the in-situ design causing for example a particular air flow around the building, wind design or specific heat release at night.
In common \gls{QD} problem domains, the genome-phenotype (and phenotype-fitness) mapping does not suffice to describe the representation, as we are interested not only in the fitness metric derived by the behavior in an environment, but by other aspects of the behavior as well.
\gls{QD} is often used to evolve robot controller behavior strategies or, in our case, a three-dimensional voxel mesh that acts on a flow.
To this end, the definition of a representation in \gls{QD} is expanded to include mapping the phenotype to the extended phenotype~\citep{dawkins1982extended}, which also includes a solution's ``behavior'' in an environment~\citep{hagg2021discovering}.

\begin{figure}
	\centering
	\includegraphics[width=0.7\linewidth]{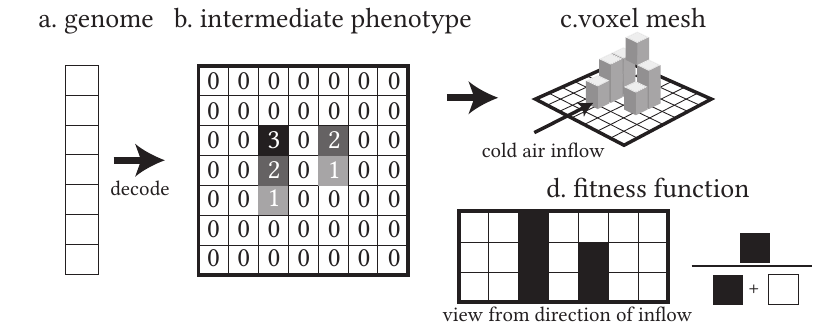}
	\caption{An encoding defines how a genome (a) is decoded into the phenotype (b, c). Illustrated example contains less grid cells than the actual phenotype used in this article. The fitness of a solution is calculated by the relative number of built cells in the direction of the cold air inflow (d).}
	\label{fig:mesh}
\end{figure}

The phenotype in this work is defined as a $11\times14$ grid of $3\times3$ meter cells (Figure~\ref{fig:mesh}b) which contain height values.
The extended phenotype is a mesh based on the grid where each cell contains a mass of either 0, 3, 6 or 9 meters high (Figure~\ref{fig:mesh}c), situated in a flow.
The problem domain is inspired by a real world construction project with urban planning constraints.
The dimensionality of the phenotype is 154.
The computationally expensive \gls{CFD} that would be used in city planning and architectural design optimization is replaced using a very simple surrogate for the experiments in this work: the relative built up area facing the main direction of cold air inflow (Figure~\ref{fig:mesh}d). 
Minimizing this area is a proxy for minimal air flow impact, good enough to produce realistic and explainable results, while keeping computational demand low. 
The most straightforward representation of massing design voxel meshes would be an array of height values.
However, a plethora of more complex representations exist.
We give an overview of those that are commonly used in shape optimization problems similar the three-dimensional voxel domain used in this article.

In the rest of this section, the \gls{QD} algorithm and a number of common representations are described, including their mutation operators. 
Finally, a multi-encoding \gls{QD} algorithm is introduced that explores the solution space using different representations simultaneously.

\subsection{MAP-Elites}
\label{sec:methods:mapelites}

One of the first \gls{QD} algorithms, \gls{MAP-Elites}~\citep{Cully2015}, uses a multidimensional grid-like archive as a phenotypic niching space (see Algorithm\ref{alg:ME}).
Its dimensions are defined by phenotypic features, which in our case are the \textit{total built area} and the \textit{number of separate buildings}.
After initializing a population of random genomes $\mathcal{X}$, their phenotypes $\mathcal{P}$ are derived through the decoding function.
Based on these phenotypes, the quality (fitness) and features of the individuals can be derived.
The individual genomes are placed in the archive $\mathcal{A}$.
New individuals are only accepted if they are better than the individual that already occupies the niche, or if the niche is empty.
The archive serves as a population from which random parents are drawn. 
The parents $\mathcal{X}_p$ are perturbed using a mutation operator to produce children $\mathcal{X}_c$.
The children $\mathcal{X}_c$ then replace those individuals in the archive that belong to the same niche, if the old niche elites perform worse or the niche is empty.
This procedure is repeated for a certain maximum number of generations, or a stopping criterion is reached.

\begin{algorithm}
	\caption{MAP-Elites algorithm}
	\label{alg:ME}
	\begin{algorithmic}
		\State $\mathcal{X}, \mathcal{A} \gets Initialize()$ \Comment{Initialize genomes and archive}
		\Procedure{map-elites}{$\mathcal{X}, \mathcal{A}$}
		\State $\mathcal{P} \gets \mathcal{D}(\mathcal{X})$ \Comment{Decode genomes into phenotypes}
		\State $\mathbf{f, p} \gets Fitness(\mathcal{P})$ \Comment{Get features and performance}
		\State $\mathcal{A} \gets Replace(\mathcal{A},\mathcal{X},\mathbf{f},\mathbf{p})$ \Comment{Replace niches}
		\While {$gens < maxGens$}
		\State $\mathcal{X}_p \gets Random(\mathcal{A})$ \Comment{Select random parents}
		\State $\mathcal{X}_c \gets Perturb(\mathcal{X}_p)$ \Comment{Perturb parents}
		\State $\mathcal{P}_c \gets \mathcal{D}(\mathcal{X}_c)$
		\State $\mathbf{f, p} \gets Fitness(\mathcal{P}_c)$ 
		\State $\mathcal{A} \gets Replace(\mathcal{A},\mathcal{X}_c,\mathbf{f},\mathbf{p})$ 
		\EndWhile
		\EndProcedure
	\end{algorithmic}
\end{algorithm}

\subsection{Direct Encodings}
\label{sec:methods:direct}

\begin{figure}[htb]
	\centering
	\includegraphics[width=0.7\linewidth,page=1]{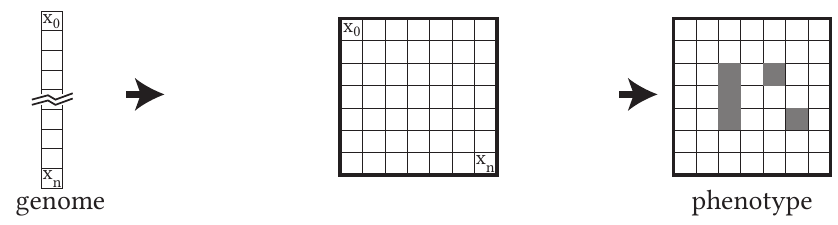}
	\caption{\textit{Direct encoding:} the height values of the cells are directly encoded into the genome.}
	\label{fig:enc:direct}
\end{figure}

The simplest encoding type performs a one-to-one mapping of the genome to the phenotype (Figure~\ref{fig:enc:direct}).
The decoding function that translates genomes into phenotypes is defined as follows:
\begin{equation}
\mathcal{D}: \mathbf{p} = \mathbf{x}, \mathbf{x} \in \mathcal{X}, \mathbf{p} \in \mathcal{P}
\label{eq:direct}
\end{equation}
This encoding has been the first to be widely adapted in \gls{EA}~\citep{rechenberg1973evolutionsstrategie,schwefel1977evolutionsstrategien}.
For the voxel mesh problem, each height value of the voxel grid would be represented by a number (usually real-valued but it can be binary as well) and be given its own locus in the genome.
The resulting high dimensionality of the genome makes the search process cumbersome and slow and is not expected to easily produce high-quality solution sets.
Technically, we could expect this encoding to be able to find a very diverse set of solutions, as all potential phenotypes can be reached. 
However, this is only reasonable as long as all parts of the search space are reachable and do not have to cross through invalid regions. 
The search process could also take an insurmountable amount of time in this high-dimensional space.

The height of every grid cell is directly encoded.
The dimensionality of this encoding therefore equals the dimensionality of the phenotype, $d = 154$.
Solutions are mutated with a small probability $p$.
A mutation increases or decreases a gene's value by 1, only allowing values from the minimum to maximum height.

\subsection{Dictionary-based}
\label{sec:methods:dict}
\begin{figure}[htb]
	\centering
	\includegraphics[width=0.7\linewidth,page=2]{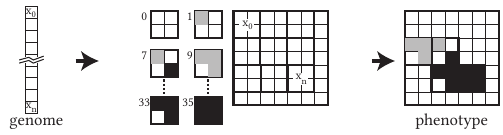}
	\caption{\textit{Dictionary encoding:} grid cells are grouped and for every group all buildings blocks are precalculated. The genome encodes which cell group gets assigned which building block.}
	\label{fig:enc:dictionary}	
\end{figure}

In order to reduce the dimensionality of the search space, a dictionary of building blocks can be used to generate solutions (Figure~\ref{fig:enc:dictionary}).
This approach is often used in floorplan layout design optimization problems~\citep{modrak2021calibration}.
The decoding function is defined as:
\begin{equation}
\mathcal{D}: \textbf{p} = \mathbf{DICT}(\textbf{x}), \textbf{x} \in \mathbb{N}
\end{equation} 
$\mathbf{DICT}$ is a dictionary of building blocks from which a vector of elements $\textbf{p}$ is selected based on an integer-valued genome $\textbf{x}$.
This method was used for city layout design optimization~\citep{xu2019revealing} where the phenotype consists of cells whose state is determined by an element from the dictionary.
The dimensionality of the search space is determined by the size of the dictionary elements.

The dictionary contains building blocks of size $n$ by $m$, with all possible states of voxel fillings that do not contain floating voxels.
It can therefore create exactly the same shapes as the direct encoding.
The mutation operator can be used to determine which state transitions are allowed.
It is defined such that, with a small probability $p$, a random state transition of the gene is performed according to the allowed rules in the dictionary.
A mutation will only change the building block such that the cells in the old and new states have a Manhattan distance of 5, to provide a more conservative mutation operator.

\subsection{Parametric Encodings}
\label{sec:methods:parametric}

Parametric encodings are common in engineering design, as they are easy to understand and allow a high degree of control.
Spline encodings are an example where key points of splines can be moved, which allows parameters to only move local shape features~\citep{sobieczky1999parametric}.
Other parametric encodings include sets of coordinates that determine the locations or sizes of a set of cubes~\citep{kaushik2013evolutionary}.
As a trade-off to the high level of control, these encodings are often quite restricting in terms of the diversity of solutions that can be produced.
To counter this, more generalized parametric encodings have been developed, such as the additive or subtractive parametric encoding~\citep{wang2019subtractive}, by allowing the user to predetermine global features of the design, such as the number of buildings~\citep{wang2020enabling}.
The dimensionality of the search space of such encodings is generally lower than a direct encoding, but the diversity of the resulting solution set might suffer.

\begin{figure}[htb]
	\centering
	\includegraphics[width=0.7\linewidth,page=3]{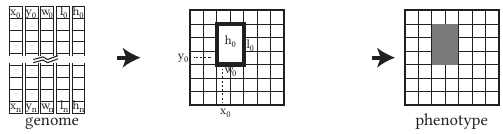}
	\caption{\textit{Parametric encoding:} the genome encodes the position and length sizes of a fixed number of rectangles.}
	\label{fig:enc:parametric}	
\end{figure}

Many parameterizations are possible but for the problem domain in this article, the following definition is used.
The decoding process $\mathcal{D}$ generates $n$ rectangles, defined by four parameters: the $x$ and $y$ position of the rectangle, having length $l$ and width $w$ (Figure~\ref{fig:enc:parametric}).
The mutation operator, with a small probability $p$, adds an integer value according to the values of a rounded up normal distribution with standard deviation $\sigma$.
The rectangles are not allowed to be placed outside of the buildable area.
It is ensured that $w$ and $l$ are non-negative.

\subsection{Compositional Pattern Producing Networks}
\label{sec:methods:cppn}

\begin{figure}[htb]
	\centering
	\includegraphics[width=0.7\linewidth,page=4]{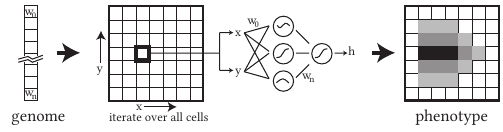}
	\caption{\textit{\gls{CPPN} encoding:} The weights and activation functions of a fixed neural network are encoded by the genome.}
	\label{fig:enc:cppn}
\end{figure}

\Glspl{CPPN} were introduced as a way to directly insert geometrical abstractions into developmental encodings~\citep{stanley2006exploiting}. 
The representation was used on voxel optimization problems~\citep{auerbach2010evolving,barthet2022open}.
\glspl{CPPN} are compositions of a variety of mathematical functions that produce a phenotype by simple cell-wise query of a substrate.
A substrate assigns coordinates to the phenotype, which can be a grid or locations of motors in a robotical system.
The encoding has been used to create two- and three-dimensional pixel/voxel grids, for example to generate three-dimensional objects~\citep{clune2011evolving}.
The decoding is defined as follows:
\begin{equation}
\mathcal{D}: p(x,y) = G(x,y), \text{where $G$ a neural graph}
\label{eq:cppn}
\end{equation}
The neural graph $G$ returns a value for each pixel or voxel in the phenotype (Figure~\ref{fig:enc:cppn}).
\glspl{CPPN} are usually evolved using \gls{NEAT}, a neuroevolution method that evolves the structure of graph objects~\citep{stanley2002evolving}, although any neuroevolution algorithm can be used.

In the case of a grid-wise phenotype, the ($x$,$y$) coordinates of the grid cells serve as a query input and the height of the building block is determined by the \gls{CPPN}.
\gls{NEAT} grows neural graphs and is therefore not appropriate for the analysis in this article, due to the variable dimensionality of the search space.
Instead of using \gls{NEAT}, the dimensionality of the search space is controlled by using a fixed architecture for the graph, containing one or two hidden layers and a preconfigured number of hidden neurons.
The genome encodes the weights of this fixed neural network graph consisting of $l$ layers and $n$ neurons that have an activation function randomly appointed from Gaussian, tanh, sigmoid, sine, cosine, constant zero, constant one and a step function.
With a small probability $p$, the mutation operator randomly selects a new activation function, or adds a small value to the weight, drawn from a normal distribution with standard deviation $\sigma$.

\subsection{Cellular Automata}
\label{sec:methods:ca}
\begin{figure}[htb]
	\centering
	\includegraphics[width=0.7\linewidth,page=5]{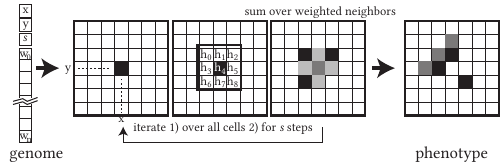}
	\caption{\textit{\gls{CA} encoding:} the weights of the influence of neighboring cells in a Moore neighborhood are encoded in the genome. The phenotype is developed in a fixed number of iterations.}
	\label{fig:enc:ca}	
\end{figure}

Developmental encodings like \gls{CA}, introduced by Von Neumann~\citep{neumann1966theory}, consist of a rule set and are developed over a number of iterations. 
This time factor determines how long the state an object can develop, based on neighborhood rules.
In \gls{CA}, the cells in homogeneous lattice grids can take on a given finite number of states, which change according to simple rules in relation to a cell neighborhood. 
Commonly used is the Moore neighborhood~\citep{moore1962machine}, which consists of the state of a cell and all its eight neighbors in the previous time step.
The extended version allows taking neighbors that are more than one Chebyshev distance away from the cell~\cite{nayak2014cellular}.
The decoding process consists of iterating over all cells in the phenotype for $t$ iterations, where $t$ represents the length of the development process (Figure~\ref{fig:enc:ca}).
In each iteration, for each cell, the value of the cell in the next iteration is calculated as follows, depending on the old values of it and its neighbors:
\begin{equation}
\mathcal{D}: p(x,y)^{t+1} = \Sigma^{j+1}_{j-1}\Sigma^{i+1}_{i-1}(w_{i,j} \cdot p(i,j)^{t})
\label{eq:ca}
\end{equation}
In basic \gls{CA}, the weights $w_{i,j}$ are often binary rules that check whether a neighbor is either on or off, and determines the new value in a cell accordingly.
The most well-known variant is Conway's Game of Life~\citep{gardner1970fantastic} that includes rules about how many neighbor cells are alive or dead.
Continuous-valued weights have been used in neural \gls{CA} approaches~\citep{nichele2017neat,mordvintsev2020growing}.

Emerging patterns are difficult to predict, as they evolve over time and are determined by local interactions. 
Two main lines of research exist: employing \gls{CA} either in goal-directed ways or in an open-ended explorative manner~\citep{herr2016cellular}. 
Predicting the effect of transition rule sets are hard, but this might be beneficial to the exploratory character of algorithms in \gls{QD}.

This work uses a simple neural cellular automata setup.
First, a seed at position $(x,y)$, encoded in the genome, is set to 1.
In order to fully express the phenotype, a cell's height values are determined based on an iterative developmental procedure as follows.
An extended Moore neighborhood of size $n \times n, n \ge 3$, contains weights that are used in Equation~\ref{eq:ca} to calculate a cell's value in the next time step.
This equation is applied to all cells for a number of time steps $t$.
The mutation operator, with a small probability $p$ moves the seed position's $x$ and $y$ coordinates by 1. 
The weights of the Moore neighborhood mask are changed according to a normal distribution.

\subsection{Other Encodings}
Other encodings have been applied to similar problem spaces, but are out of the scope of this work.
The most prominently used other encoding are procedural grammar-based, which can use chains or trees of commands from \gls{CAD}-programs to create objects~\citep{wonka2003instant}.
Graph-based encodings can also be used to plan a design based on a predefined set of \gls{CAD} building blocks~\citep{keshavarzi2021genfloor}.
This article does not consider these software-driven encodings due to their proprietary nature and licensing models but could be considered in a more extensive analysis in the future.

Deep learning has created major breakthroughs in generative design.
Data-driven generative models such as \gls{VAE}~\citep{kingma2013VAE} can be used to learn an encoding from data~\citep{gaier2020discovering}. 
Although these data-driven generative models can move knowledge from later into earlier design phases, they depend on the availability of a large amount of training data, which is usually not publicly available.
Especially smaller architectural firms, communities and projects can therefore not rely on such techniques and can become more dependent on privatized models.
However, \gls{QD} can be used as a precomputation method to generate diverse training data sets.
The generative model then produces high quality solutions~\citep{gaier2020discovering} that can also adhere to design space constraints~\citep{bentley2022evolving}.
Precomputation can solve the problem of large computation time usually seen in evolutionary design systems~\citep{janssen2022mobius}.

Because of the aforementioned reasons, a large interest remains in non-data-driven and openly available representation techniques.
It is therefore of interest to compare which of these techniques can generate the most diverse and high-quality solution set.

\begin{figure*}[htb]
	\centering
	\includegraphics[width=1.0\linewidth]{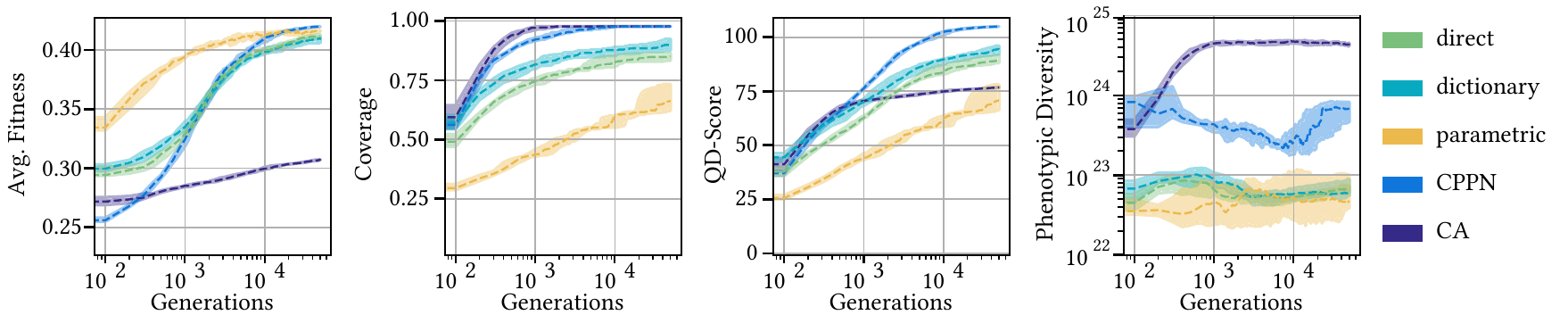}
	\caption{Evolution of all representations. Included is our approach that combines multiple encodings in MAP-Elites. Experiments are replicated 10 times. Shown are the median (dashed line) and 25\%/75\% percentiles.}
	\label{fig:eval:evolution}
\end{figure*}

\subsection{Multi-encoding QD}
\label{sec:methods:meqd}

We also evaluate whether making use of mixed representations in QD advances the diversity of the archive.
Multi-encoding \gls{QD} contains differently encoded species that compete on a phenotypic level.
Although classical speciation is built around the fact that individuals can only breed with members of their own species through crossover, we refrain from using the crossover mechanism entirely in this preliminary analysis.
Due to the fact that MAP-Elites does not use a crossover operation, using multiple encodings in MAP-Elites is straightforward.
Each encoding comes with its own mutation operator and solutions are only compared based on their phenotypic features and fitness.
The initial population contains equally many individuals from all encodings.

\section{Evaluation}
To answer the research questions that were posed in Section~\ref{sec:introduction}, the direct and indirect representations described in Sections \ref{sec:methods:direct}--\ref{sec:methods:ca} are compared.
In order to do this, \gls{MAP-Elites} is run ten times for every representation for 50,000 generations from an initial population of 100 individuals, creating ten children in each generation.
The surrogate \textit{fitness} function that was described in Section~\ref{sec:methods} is used to determine the quality of solutions. 
We measure how well the archive is filled by using the \textit{coverage} metric, the percentage of bins that are filled.
The \textit{\gls{QD}-score}~\citep{pugh2015confronting} sums up the fitness values over all cells in the archive, and has become a standard metric to measure the performance of \gls{QD} algorithms that use a fixed grid.
The score, however, depends on the feature space that is used in the problem domain and does not capture the full diversity of the solutions.
To measure \textit{phenotypic diversity}, the sum of all pair-wise distances of phenotypes in the archive is calculated.
By using the $L0.1$-norm to measure the distance between individuals, the diversity metric is made more robust for high-dimensional spaces~\citep{wang2016diversity}.

\begin{figure*}[h!tb]
	\includegraphics[width=1.0\linewidth]{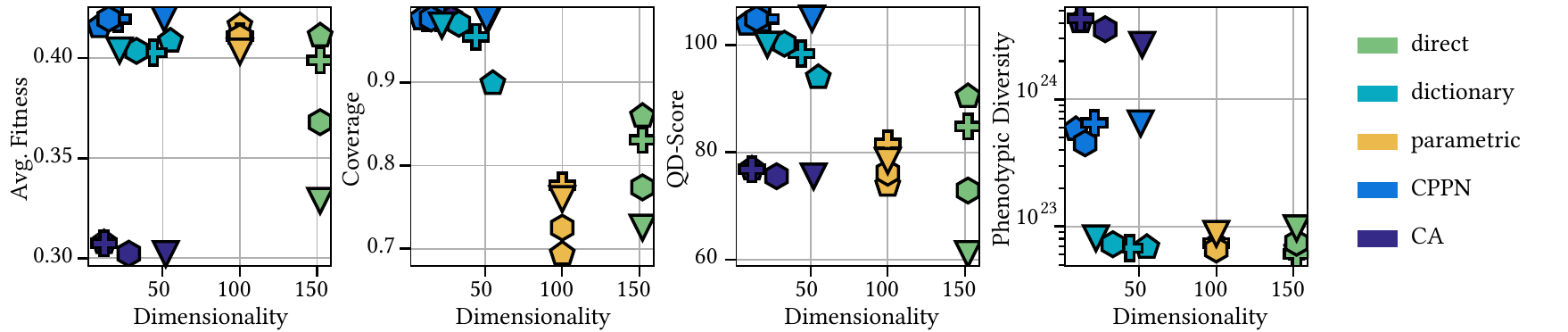}
	\caption{Relationships between four metrics and the encodings' dimensionality. Different shapes denote different parameterizations for each encoding. Please refer to Section E in the Appendix to retrieve the values of the parameters.}
	\label{fig:2by3}
\end{figure*}

Furthermore, the representations' dimensionality is measured by counting the number of degrees of freedom the representation offers.
We also evaluate, whether representations find similar solutions or distinct regions in the solution space.
Finally, representations are combined, as described in Section \ref{sec:methods:meqd}, to determine what effect this has on \gls{QD}'s performance.
Two-sampled t-tests are used to determine the significance between fitness, coverage, \gls{QD}-score and phenotypic diversity of the solution sets.

\subsection{Hyperparameters}

The various hyperparameters of the encodings are determined by parameter sweeps using a grid search.
For each encoding, \gls{MAP-Elites} is run ten times for every representation for 25,000 generations with an initial population of 100, creating ten children in each generation.
The average fitness and phenotypic diversity were ordered in fronts according to Pareto-dominance.
Then, starting from the first Pareto front, the best four hyperparameter configurations were selected.
The hyperparameter values can be taken from the code repository.

\begin{figure*}[h!tb]
	\includegraphics[width=1.0\linewidth]{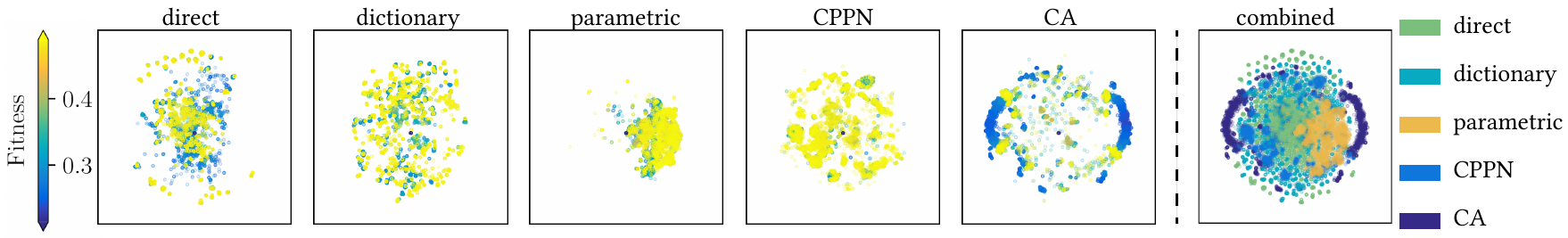}
	\caption{Visualization of phenotypic space of all encodings using t-SNE. Encodings are color coded by fitness and combined on the right to show what parts of the phenotype space they reach.}
	\label{fig:tsne}
\end{figure*}

\subsection{Quality and Diversity of Representations}
\label{sec:eval:qd}

The evolution for each representation is shown in Figure~\ref{fig:eval:evolution}.
The quality of the representations at initialization is very different.
During the first 20,000 generations, the fitness of the parametric encoding outperforms all other representations.
After this point in the runs the \gls{CPPN} overtakes the parametric encoding to find the highest quality solutions.
The performance of the direct and dictionary-based encodings is similar, which is expected, as they can express the same shapes.
They do not reach the same fitness levels as the parametric and \gls{CPPN}.
The \gls{CA}'s fitness is the lowest of all encodings.
The archive coverage starts out in essentially the opposite order: the \gls{CA} immediately outperforms the other encodings, although the \gls{CPPN} ends up at the same level.
The direct encoding only outperforms the parametric encoding in terms of coverage and underperforms compared to other indirect encodings.

The \gls{QD}-score shows a more complete story, as both the coverage and fitness are taken into account.
The \gls{CPPN}-encoding finds the highest-performing solutions over the largest part of the archive, although it takes close to a 1,000 generations before it overtakes the other encodings.
The \gls{CA}-encoding underperforms in quality but this is partially compensated by a large archive coverage.
The direct and dictionary-based encodings trail behind the \gls{CPPN} but perform better than the \gls{CA} and parametric encodings.
The phenotypic diversity is shown on the right of the figure.
The \gls{CA} and \gls{CPPN} encodings are clearly more phenotypically diverse.
The diversity between the parametric, direct and dictionary-based encoding are similar.
Encoding examples are shown in Figure~\ref{fig:teaser}.

Two-sample t-test were performed to determine the significance of results.
Please refer to the supplementary material to find all pair-wise significance tests.
Most comparisons between the encodings are statistically significant with $p < 0.05$. 

\subsection{Dimensionality of Representations}
\label{sec:eval:dim}

When plotting fitness, coverage, and QD-score in relation to the encodings' dimensionality, a pattern emerges in Figure~\ref{fig:2by3}. 
The best-performing encodings tend to have a lower dimensionality, both in fitness as well as in coverage.
The phenotypic diversity of the \gls{CPPN} and \gls{CA} encodings is much higher, both outperforming the other encodings, which perform similarly.
Some encodings are more sensitive to the hyperparameterization than others.
An example of this is the direct encoding's performance, which is most sensitive to the hyperparameters.
The dictionary-based encoding is able to reach a coverage similar to that of the \gls{CPPN} and \gls{CA}, depending on the hyperparameters used.
However, the encoding does not seem to be equally sensitive as the direct encoding, as the four points are not as far apart.
The indirect encodings seem to be mostly insensitive to the hyperparameterization, which makes them easier to use.

The fitness of the solution sets does not seem to depend much on the dimensionality.
However, both the coverage of the archive, the \gls{QD}-score, and the phenotypic diversity are higher for most lower-dimensional encodings.

\subsection{Different Regions in Phenotypic Space}
\label{sec:eval:regions}
Encodings might be able to reach different parts of the phenotypic space.
Figure \ref{fig:tsne} shows all solutions from all encodings in a two-dimensional similarity space that was calculated using \gls{t-SNE}~\citep{VanDerMaaten2008}. 
The embedding was calculated based on all solutions from all 10 replicates of all encodings.
Some of the indirect encodings reach different parts of the phenotypic space, although with mixed results. 
Although the \gls{CA} encoding finds very different solutions in the left and right of the space, they also account for mostly low-performing solutions. What stands out is that most encodings might indeed find noticeably different, often high-performing solutions.

\subsection{Multi-encoding QD}
\label{sec:eval:meqd}
Because the encodings found quite different solutions, an open question is, whether it makes sense to use multiple encodings in \gls{QD}.
The performance metrics of the single-encoding \gls{QD} runs are compared to multi-encoding \gls{QD} in Table~\ref{tbl:multi:performance}. 
Using multiple encodings, \gls{QD} is able to reach a \gls{QD}-score similar to that of the best single-encoding runs with \glspl{CPPN}. 
The phenotypic diversity is not as high though, and certainly not as high as the \gls{CA} runs, although the latter finds mostly lower-performance solutions.

\begin{table}[htb]
	\resizebox{\textwidth}{!}{
		\centering
	\begin{tabular}{l|l|l|l|l}
		& Avg. Fitness & Coverage & QD-Score & Phenotypic Diversity \\
		Direct Encoding & $ 0.411 \pm 0.004$ & $ 0.859 \pm 0.042 $ & $90.430 \pm 3.749$ & $ 6.653 \times 10^{22} \pm 1.044 \times 10^{22}$ \\
		Dictionary-based Encoding & $ 0.409 \pm 0.005$ & $ 0.899 \pm 0.038 $ & $ 93.980 \pm 2.971$ & $ 6.864 \times 10^{22} \pm 2.365 \times 10^{22} $ \\
		Parametric Encoding & $ 0.416 \pm 0.001$ & $ 0.694 \pm 0.092 $ & $ 73.911 \pm 9.650$ & $ 7.182 \times 10^{22} \pm 6.032 \times 10^{22} $ \\
		Compositional Pattern Producing Networks & $ \mathbf{0.420} \pm 0.0$ & $ \mathbf{0.977} \pm 0.0 $ & $ \mathbf{104.937} \pm 0.075$ & $ 6.465 \times 10^{23} \pm 2.603 \times 10^{23} $ \\
		Cellular Automata & $ 0.307 \pm 0.001$ & $ \mathbf{0.977} \pm 0.0 $ & $ 76.832 \pm 0.244$ & $ \mathbf{4.322 \times 10^{24} \pm 4.243 \times 10^{23}} $\\
		Multi-encoding & $0.414 \pm 0.002$ & $\mathbf{0.977} \pm 0.0$ & $103.504 \pm 0.496$ & $1.281\times 10^{23} \pm 0.625\times 10^{23}$
	\end{tabular}
}
	\caption{Performance of single- and multi-encoding \gls{QD}. Reported are the mean and $\sigma$ over 10 experiments.}
	\label{tbl:multi:performance}
\end{table}

\begin{figure}[h!tb]
	\centering
	\includegraphics[width=0.7\linewidth]{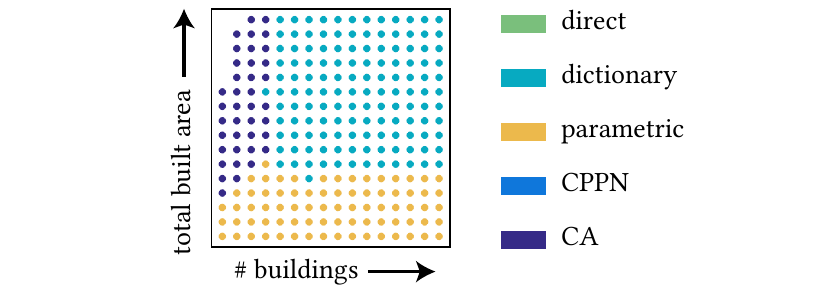}
	\caption{Example distribution of encodings in 
		an archive of multi-encoding \gls{QD}.}
	\label{fig:archive:multi}
\end{figure}

The question is, whether multi-encoding \gls{QD} relies solely on \gls{CPPN} to reach this result.
To answer this, we analyze the proportion of encodings in the population.
Figure~\ref{fig:archive:multi} shows an example archive, one where the \gls{CPPN} has died out.
The dictionary-based, parametric and \gls{CA} encodings each take up a contiguous part of the archive.

\begin{figure}[h!tb]
	\centering
	\includegraphics[width=0.7\linewidth]{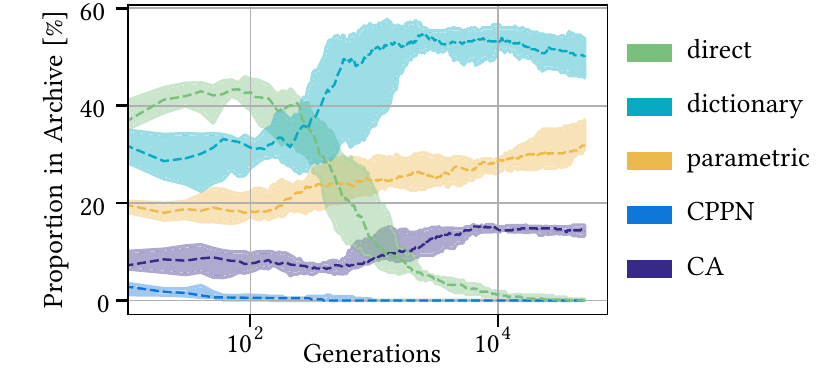}
	\caption{Proportion of encodings in multi-encoding \gls{QD}. Shown are the median (dashed line) and 25\%/75\% percentiles.}
	\label{fig:proportion_vs_generations}
\end{figure}

We investigate the proportion of the encodings in the population over all ten runs (Figure~\ref{fig:proportion_vs_generations}).
\gls{MAP-Elites} was initialized with the same number of individuals for each encoding, 20 per encoding.
However, individuals can be assigned to the same niche, leading to the removal of the weaker individual.
Through intra- or inter-species competition, the proportion of each encoding after initialization is not the same.
As can be seen from the figure, the largest portion of the archive is filled with solutions from the direct encoding in approximately the first 100 generations.
After that it gets overtaken by the dictionary-based, parametric and \gls{CA} encodings.
Interestingly enough, when we combine all encodings in a single \gls{QD} archive, the \gls{CPPN}, which performs best between all encodings, gets mostly removed from the archive after 100 generations.
This is probably due to the underperformance in early evolution.
Since we do not reinject encodings into the archive, as soon as one encoding dies out, it cannot return.

There are however two runs where \gls{CPPN} survives after 50,000 generations (see Table \ref{tbl:percentage}). 
In one of them, it takes over half of the archive.
This gives us some evidence, that we need to either protect some species that might have more difficulty in the beginning of the search, or develop other suitable methods from classical \gls{EA} to deal with the problem.

\begin{table}[tbh]
		\centering
	\begin{tabular}{l|l|l|l|l|l}		
		\textbf{\#} & CA & CPPN & Direct & Parametric & Dictionary \\
		1 & 19,2\% &    0\% &   0\% &   31.2\% &    49.6\% \\
		2 & 13.2\% & 4.4\% & 2.8\% &22.8\% &56.8\% \\
		3 & 12.8\% & 0\% &  0\% & 32.8\% & 54.4\% \\
		4 & 16.4\% & 0\% & 0\% & 40.4\% & 43.2\% \\
		5 & 15.6\% & 0\% & 0\% & 30.8\% & 53.6\% \\
		6 & 14.8\% & 0\% &  0\% & 38.0\% & 47.2\% \\
		7 & 0\% & 52.4\% & 0\% & 19.2\% & 28.4\% \\
		8 & 13.6\% & 0\% & 0.4\% & 40.8\% & 45.2\% \\
		9 & 13.6\% & 0\% & 1.6\% & 30.8\% & 54.0\%\\
		10 & 15.2\% & 0\% & 0\% & 34.4\% & 50.4\% \\
	\end{tabular}
	\caption{Proportion of encodings in multi-encoding archive. Each line is a separate run (\#).}
	\label{tbl:percentage}
\end{table}

\subsection{Discussion}
In Section~\ref{sec:eval:qd} we observed that in our experiments indirect encodings can but do not always outperform direct encodings in terms of quality, but are better at filling the archive, having a higher coverage.
Parametric encodings find high-quality solutions for only a small part of the archive early on in \gls{QD} but the \gls{QD}-score is lower than all other encodings.
In the specific domain we defined, we observed that \gls{CPPN} outperforms all other encodings, followed by the dictionary-based and direct encodings.
In Section~\ref{sec:eval:dim} we observed that low genetic dimensionality tends to produce higher coverage.
The visualization in Section~\ref{sec:eval:regions} shows that different encodings can cover distinct regions in phenotype space.
In Section~\ref{sec:eval:meqd} we observed that multi-encoding \gls{QD} outperforms all other single encodings, except that it performs similar to \gls{CPPN}, although it barely uses that encoding.

\section{Conclusion}

Contrary to the evidence shown in~\citep{tarapore2016different} we observed that indirect encodings can outperform direct encodings in \gls{QD}. 
This naturally depends heavily on whether the encoding can reach appropriate regions in phenotypic space.
Lower encoding dimensionality usually allows significantly higher archive coverage but has only a small impact on the solutions' fitness.
We observed that, although \gls{CPPN} encodings outperform others both in quality and diversity, by using multiple encodings in \gls{MAP-Elites} we can get similar results, even if \glspl{CPPN} barely survive against the other encodings.
Multi-encoded \gls{QD} is a novel approach to \gls{EA} as it allows multiple qualitatively different \textit{species} of encodings to compete based on their phenotypic (or behavioral) features.

\paragraph{Limitations}
Although using these simple surrogates instead of \gls{CFD} diminishes the realism for practical usage of produced solutions, they still allow comparing encodings while reducing the computational effort that went into this analysis.

Another limitation is that it is possible that the mutation operators themselves grant an advantage to one encoding over another, either in terms of their ability to escape local optima through larger movements in the search space, or conversely, through an ability to make more subtle movements. 

The multi-encoding approach has a weakness in its initialization due to the structured archive in \gls{MAP-Elites}.
Solutions that could evolve later on in the optimization process can be pushed out by early-evolvable encodings during initialization or in the early optimization process.
Speciation has been around for a long time in \glspl{EA} but \gls{QD} might introduce a new perspective.

\paragraph{Future Work}

A selection of common representations was compared in this work, leaving out some of the common \gls{CAD}-driven encodings, which should be included in future work.
A more rigorous and complete analysis that includes multiple categories of domains, from robotics control to shape optimization, should shed more light on the question of whether indirect or direct encodings are more appropriate in \gls{QD}.
A more indepth analysis of mutation, crossover and selection operators should be taken into consideration.
The observations in this work beg for a more theoretical analysis of representations with respect to their redundancy, scaling and locality~\citep{rothlauf2006representations}, especially how these effects might explain and improve the behavior of multi-encoding \gls{QD}.
For example, because some encodings like \gls{CPPN} evolve more slowly in the beginning, protection mechanisms or different mutation operators could be introduced.
Finally, during initialization, an unstructured archive could be used to make sure solutions are not overwritten immediately.
\\ \\
This work showed that one should be mindful about the encoding that is used when applying \gls{QD} optimization to a problem domain.
A multi-encoding approach might take more advantage of the strengths of multiple encodings, while reducing their weaknesses.
Due to the qualitative difference in encodings, we can interpret them as truly different species.

\section*{Acknowledgements}
The authors thank Rudolf Berrendorf and Javed Razzaq for their continuous development and support of the BRSU’s computing cluster.

\section*{Funding}
The computer hardware was supported by the Federal Ministry for Education and Research and by the Ministry for Innovation, Science, Research, and Technology of the state of Northrhine-Westfalia (research grant 13FH156IN6).

\bibliographystyle{apalike}
\bibliography{main} 

\clearpage
\section*{Appendix}

\subsection*{Significance Tests Fitness}
\label{appendix:significance:fitness}
\begin{figure}[thb]
	\centering
	\includegraphics[width=1\linewidth]{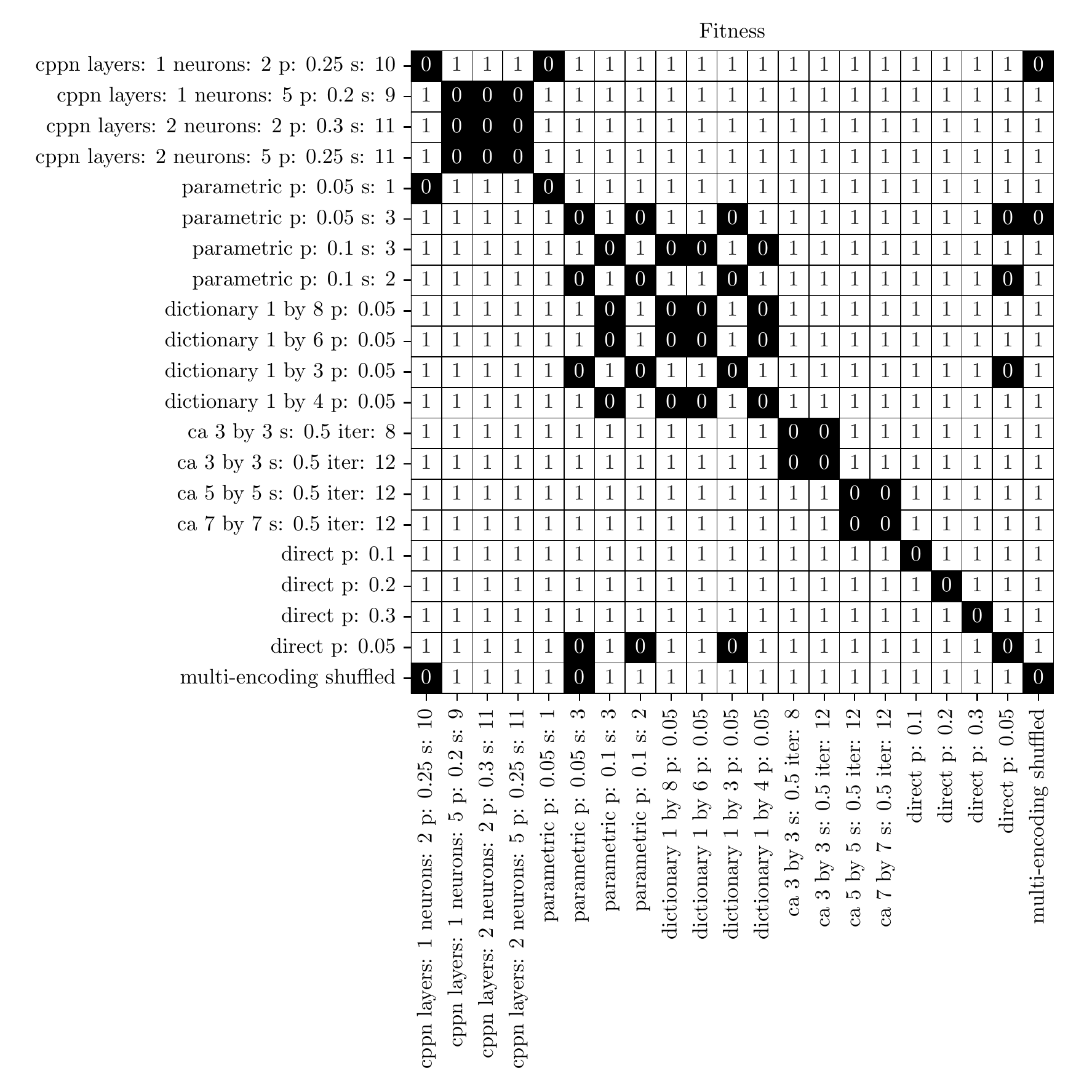}
	\caption{Two-sampled t-test results comparing pair-wise fitness values of encodings/hyperparameters, for $p < 0.05$.}
	\label{fig:ttest:fitness}
\end{figure}

\FloatBarrier
\subsection*{Significance Tests Coverage}
\label{appendix:significance:coverage}
\begin{figure}[thb]
	\centering
	\includegraphics[width=1\linewidth]{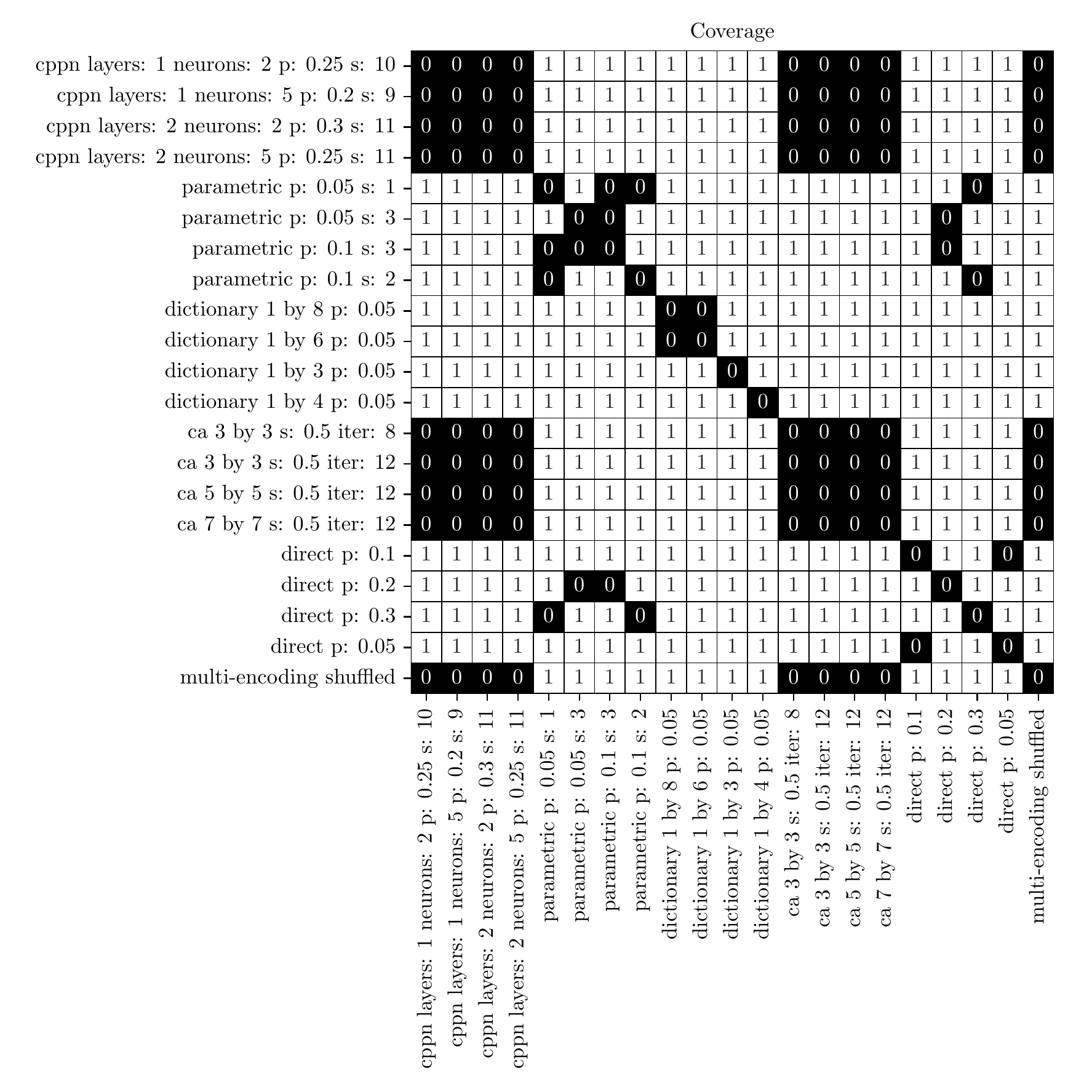}
	\caption{Two-sampled t-test results comparing pair-wise coverage values of encodings/hyperparameters, for $p < 0.05$.}
	\label{fig:ttest:coverage}
\end{figure}

\FloatBarrier
\clearpage
\subsection*{Significance Tests QD-Score}
\label{appendix:significance:qdscore}

\begin{figure}[thb]
	\centering
	\includegraphics[width=1\linewidth]{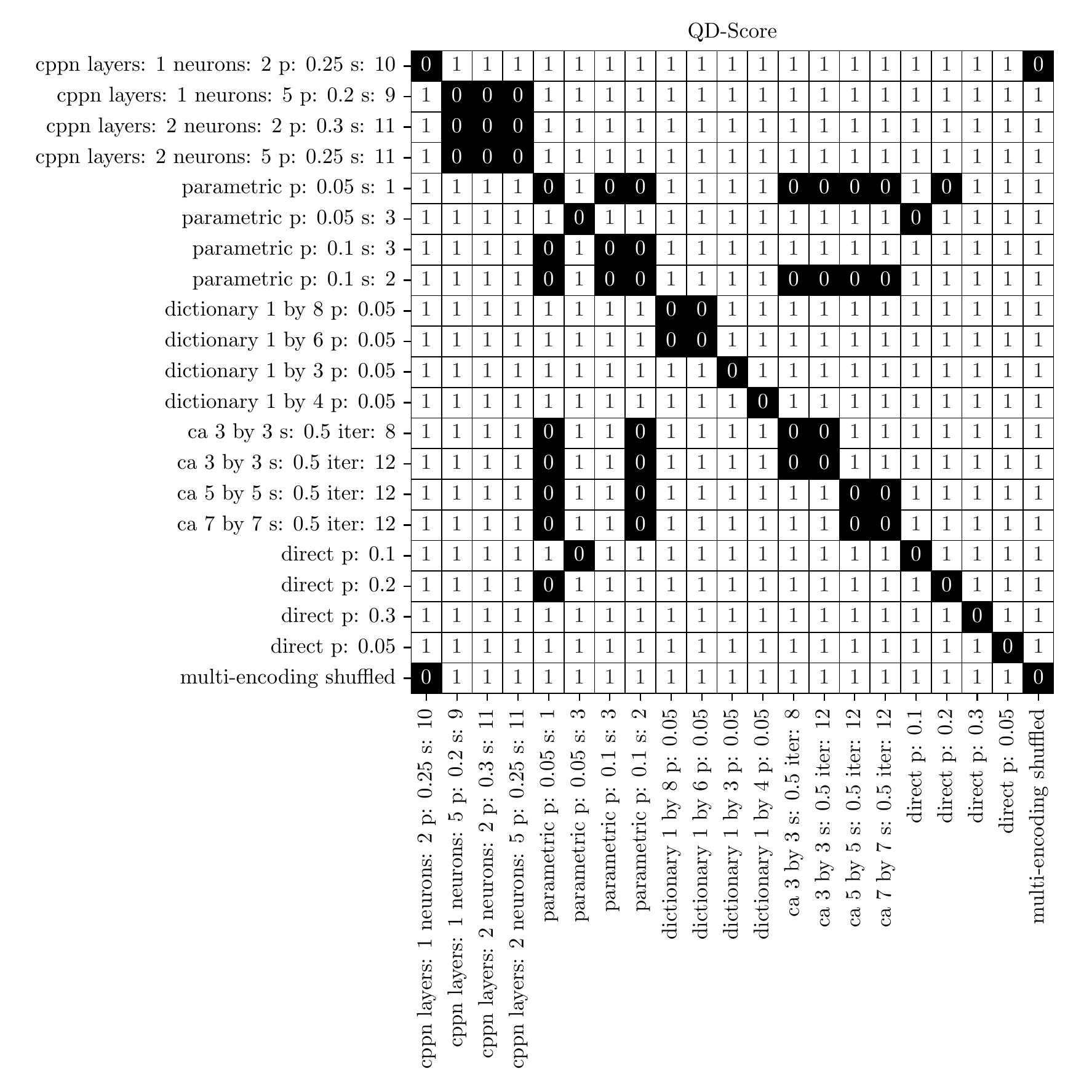}
	\caption{Two-sampled t-test results comparing pair-wise \gls{QD}-score of encodings/hyperparameters, for $p < 0.05$.}
	\label{fig:ttest:qdscore}
\end{figure}

\FloatBarrier
\subsection*{Significance Tests Phenotypic Diversity}
\label{appendix:significance:phendiv}

\begin{figure}[thb]
	\centering
	\includegraphics[width=1\linewidth]{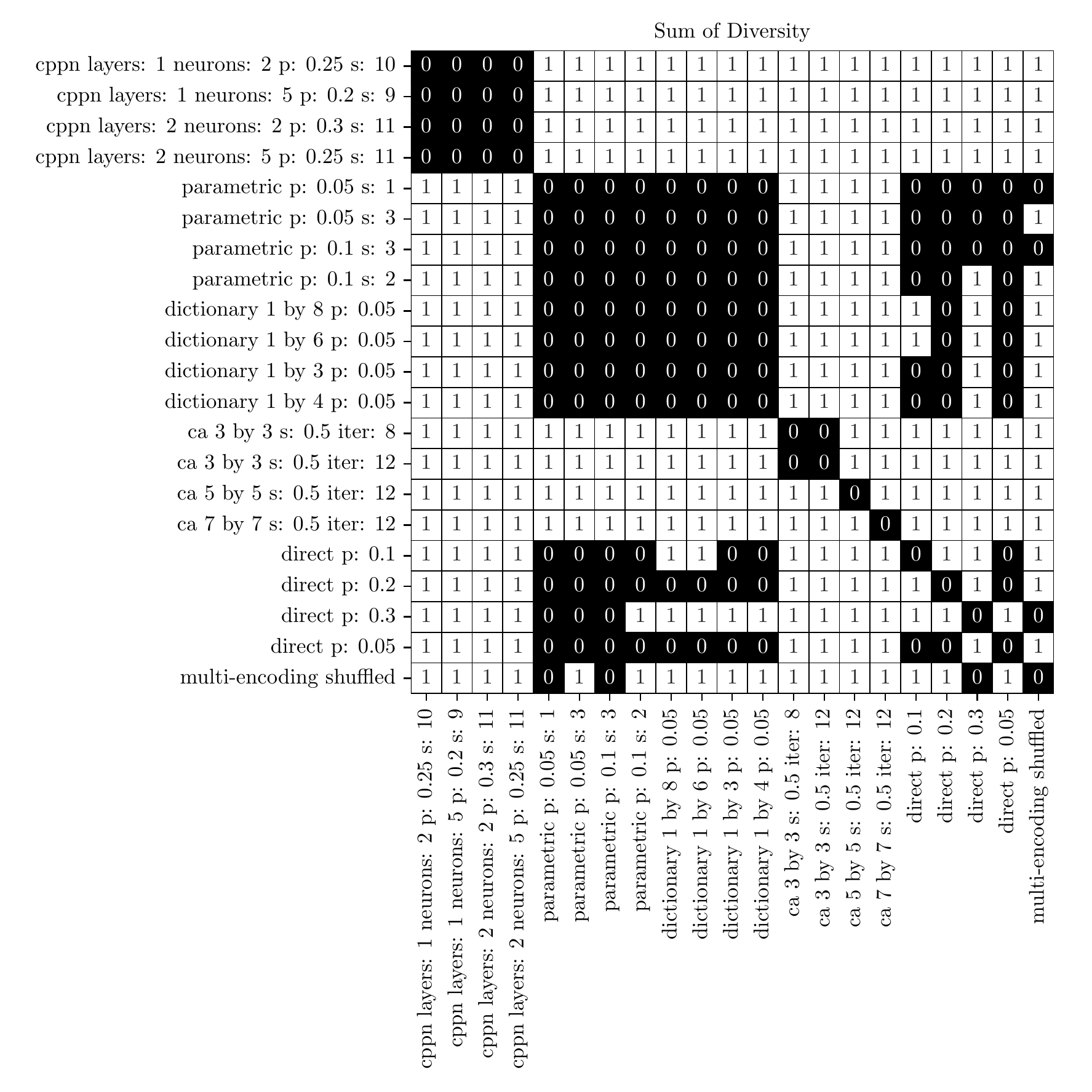}
	\caption{Two-sampled t-test results comparing pair-wise phenotypic diversity values of encodings/hyperparameters, for $p < 0.05$.}
	\label{fig:ttest:phenodiv}
\end{figure}

\FloatBarrier
\subsection*{Explanation of Symbols}
\label{appendix:explanation}

\begin{figure}[h!tb]
	\centering
	\includegraphics[width=0.7\linewidth]{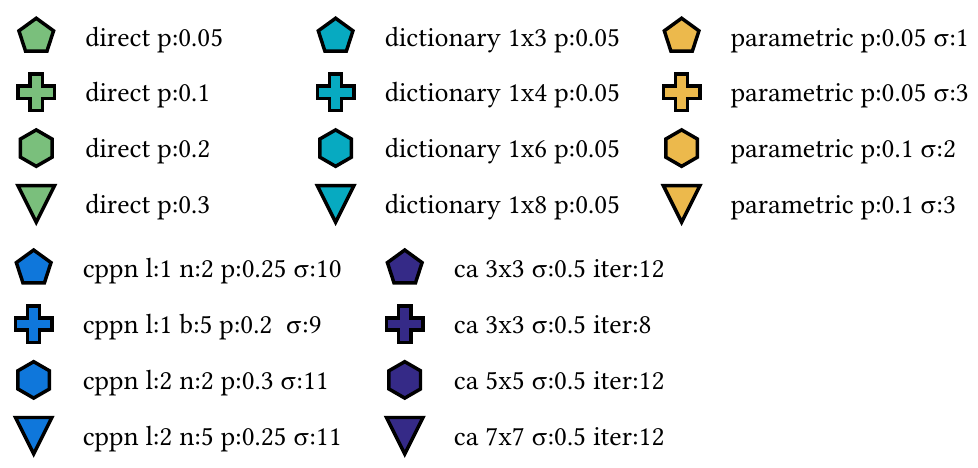}
	\caption{Explanation of different parameterizations for each encoding.}
	\label{fig:2by3explained}
\end{figure}

\clearpage
\FloatBarrier
\subsection*{More Encoding Examples}
\label{appendix:examples}

\begin{figure}[htb]
	\centering
	\includegraphics[width=0.5\linewidth]{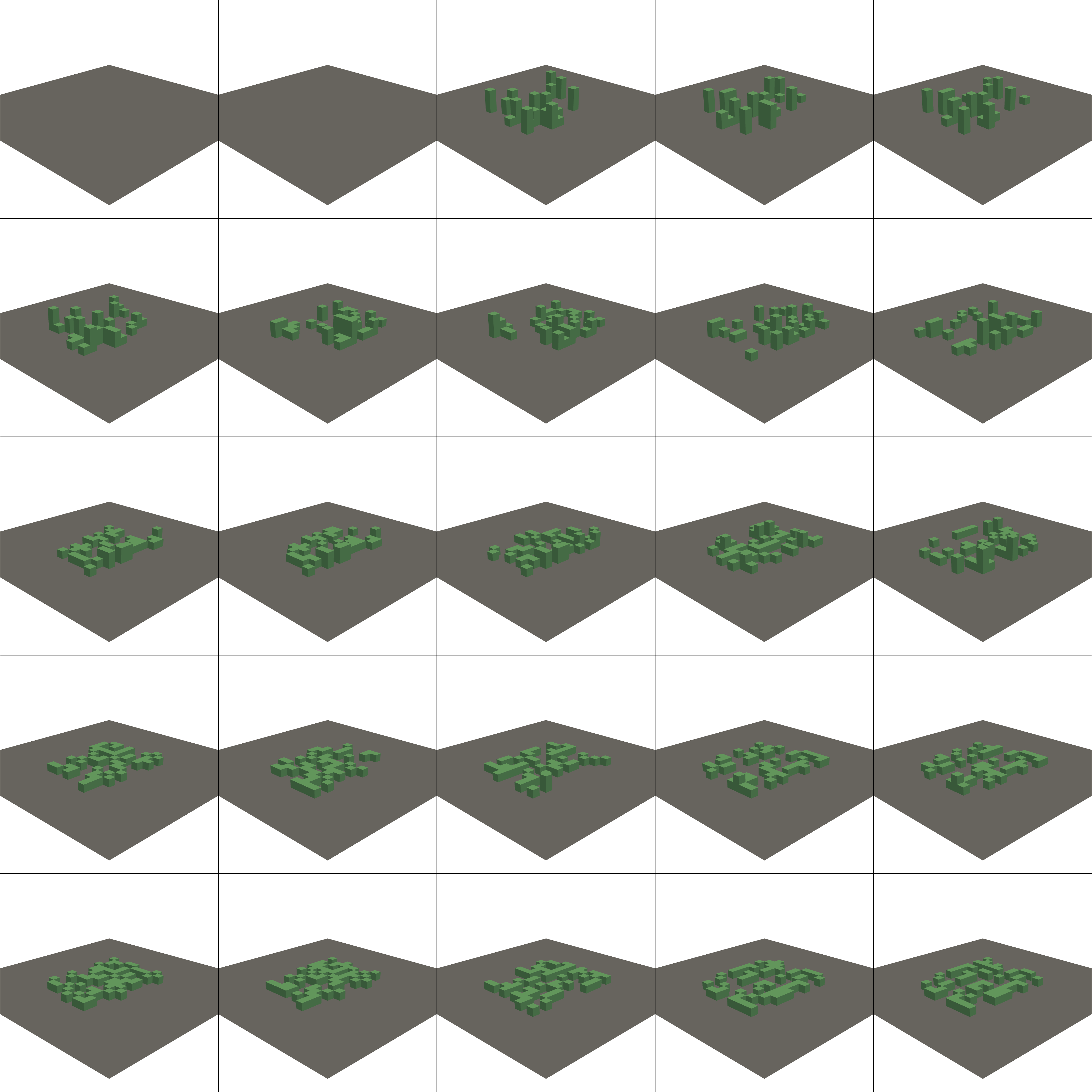}
	\caption{Randomly selected archive with 25 representative solutions using direct encoding.}
	\label{fig:examples:direct}
\end{figure}

\begin{figure}[htb]
	\centering
	\includegraphics[width=0.5\linewidth]{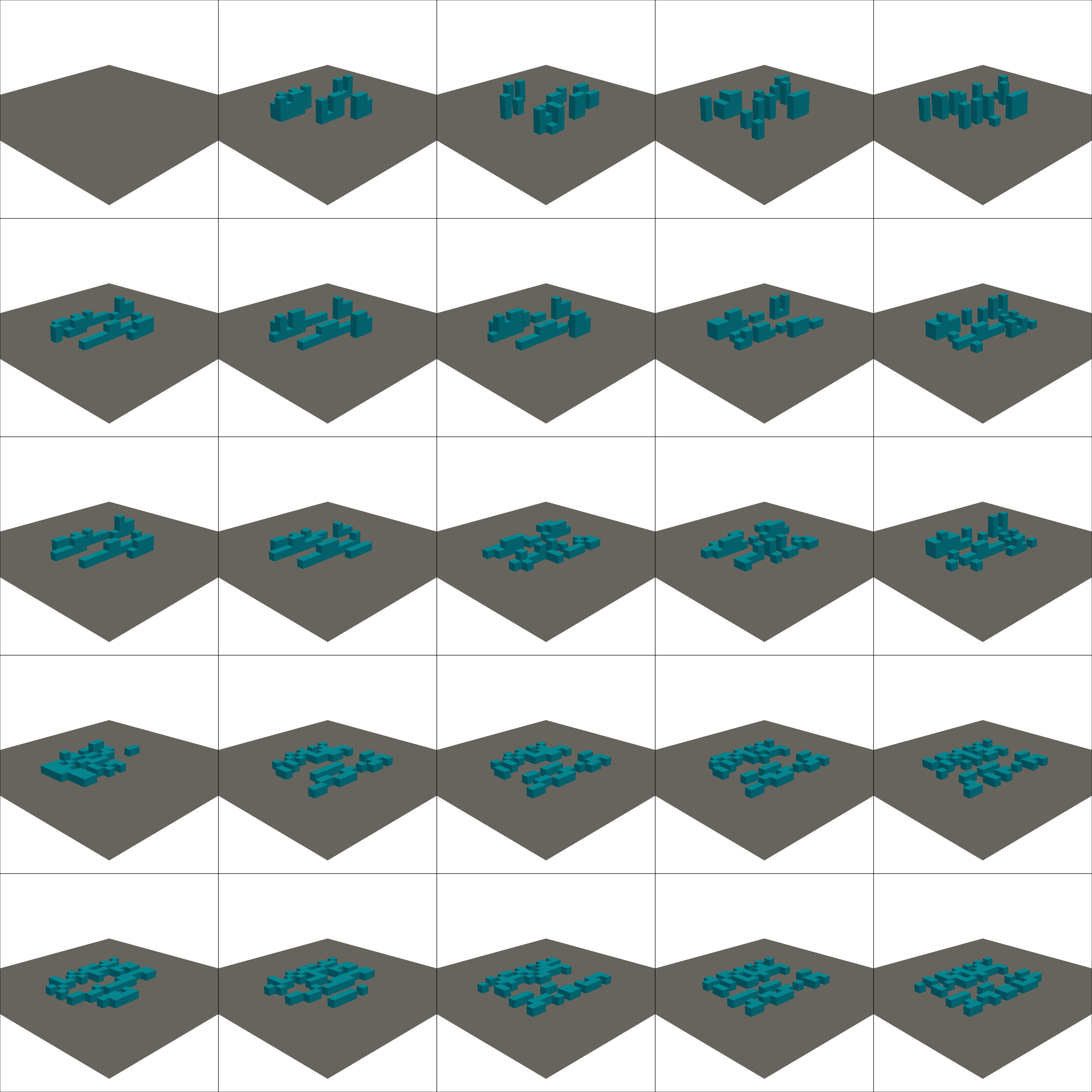}
	\caption{Randomly selected archive with 25 representative solutions using dictionary encoding.}
	\label{fig:examples:rulebased}
\end{figure}

\begin{figure}[htb]
	\centering
	\includegraphics[width=0.5\linewidth]{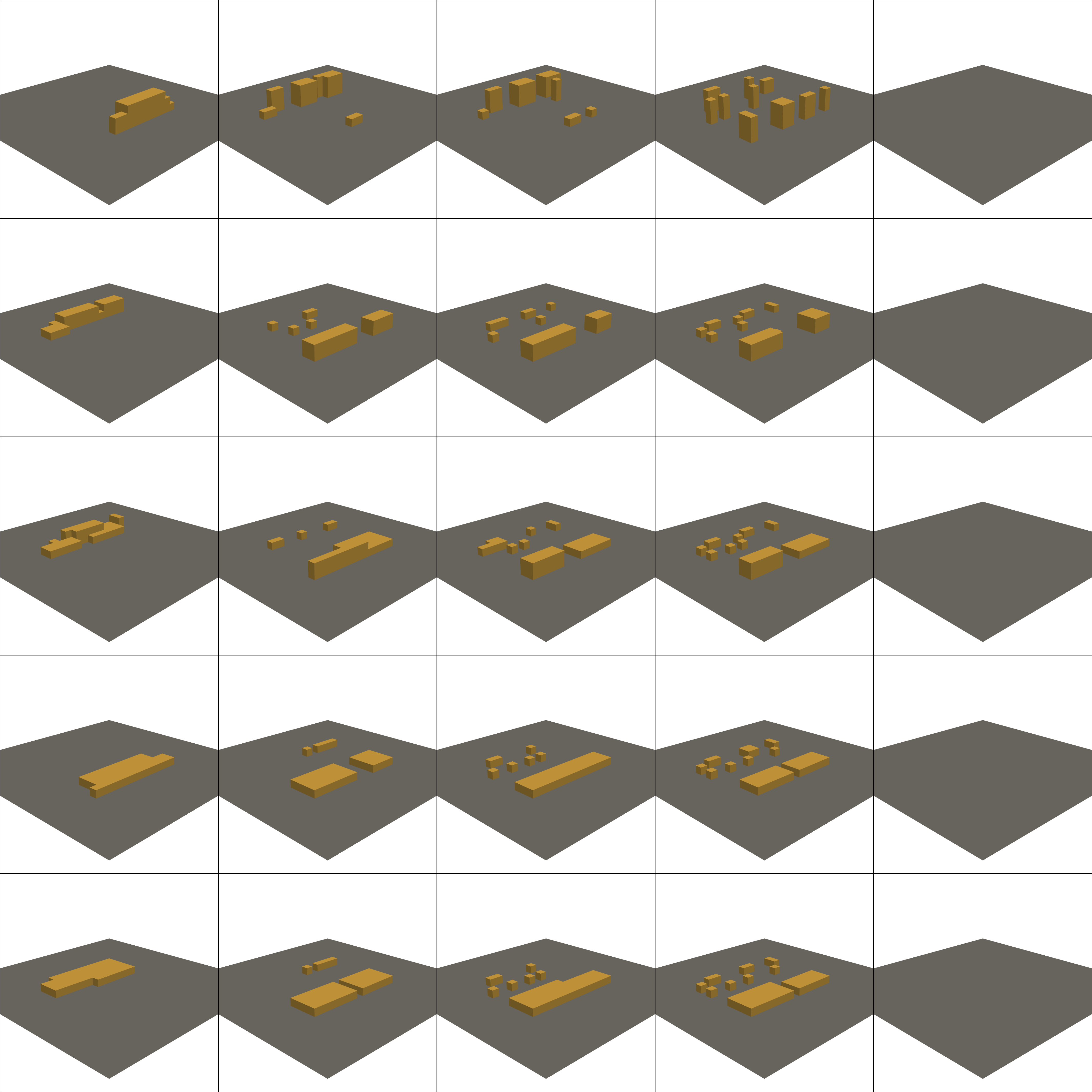}
	\caption{Randomly selected archive with 25 representative solutions using parametric encoding.}
	\label{fig:examples:parametric}
\end{figure}

\begin{figure}[htb]
	\centering
	\includegraphics[width=0.5\linewidth]{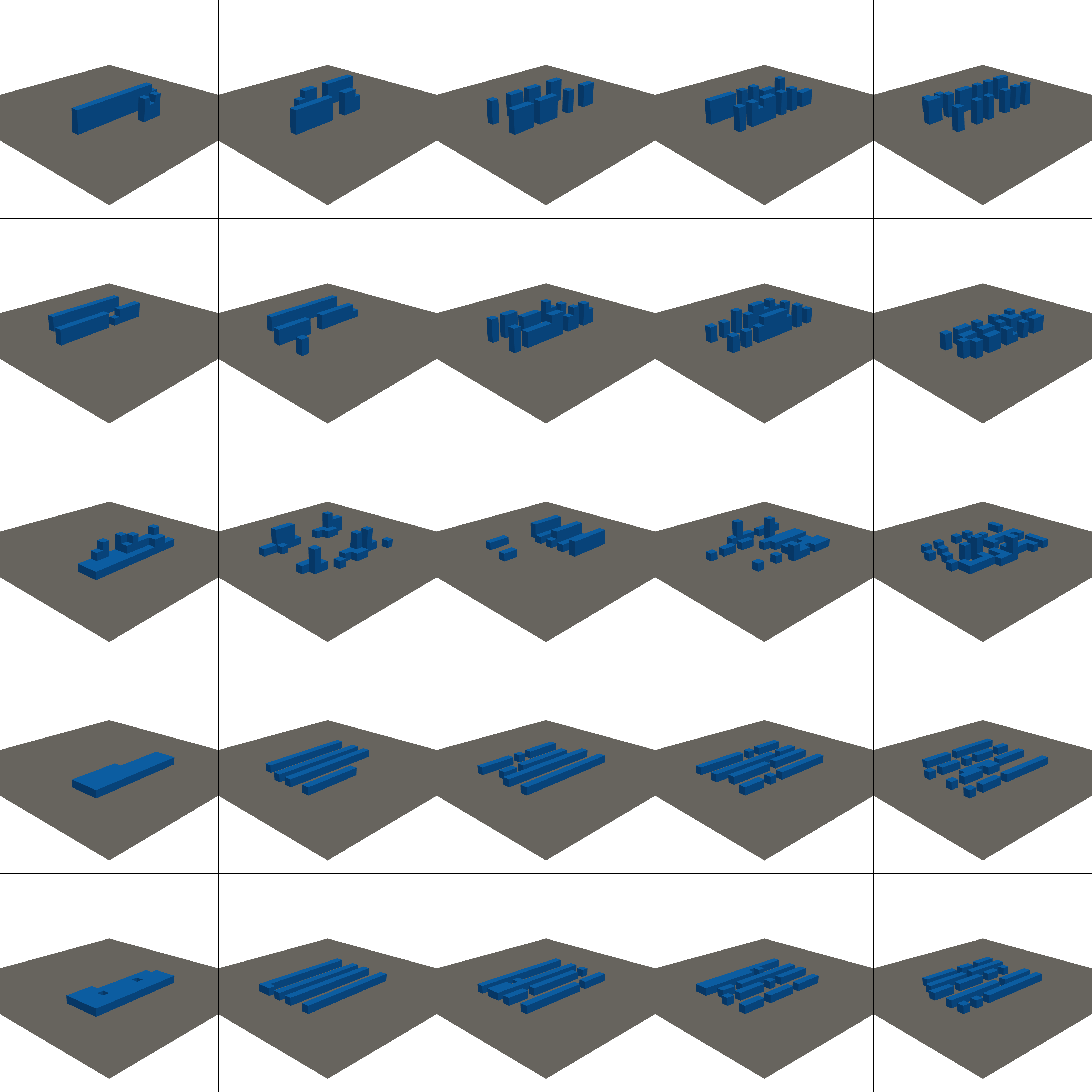}
	\caption{Randomly selected archive with 25 representative solutions using CPPN encoding.}
	\label{fig:examples:cppn}
\end{figure}

\begin{figure}[htb]
	\centering
	\includegraphics[width=0.5\linewidth]{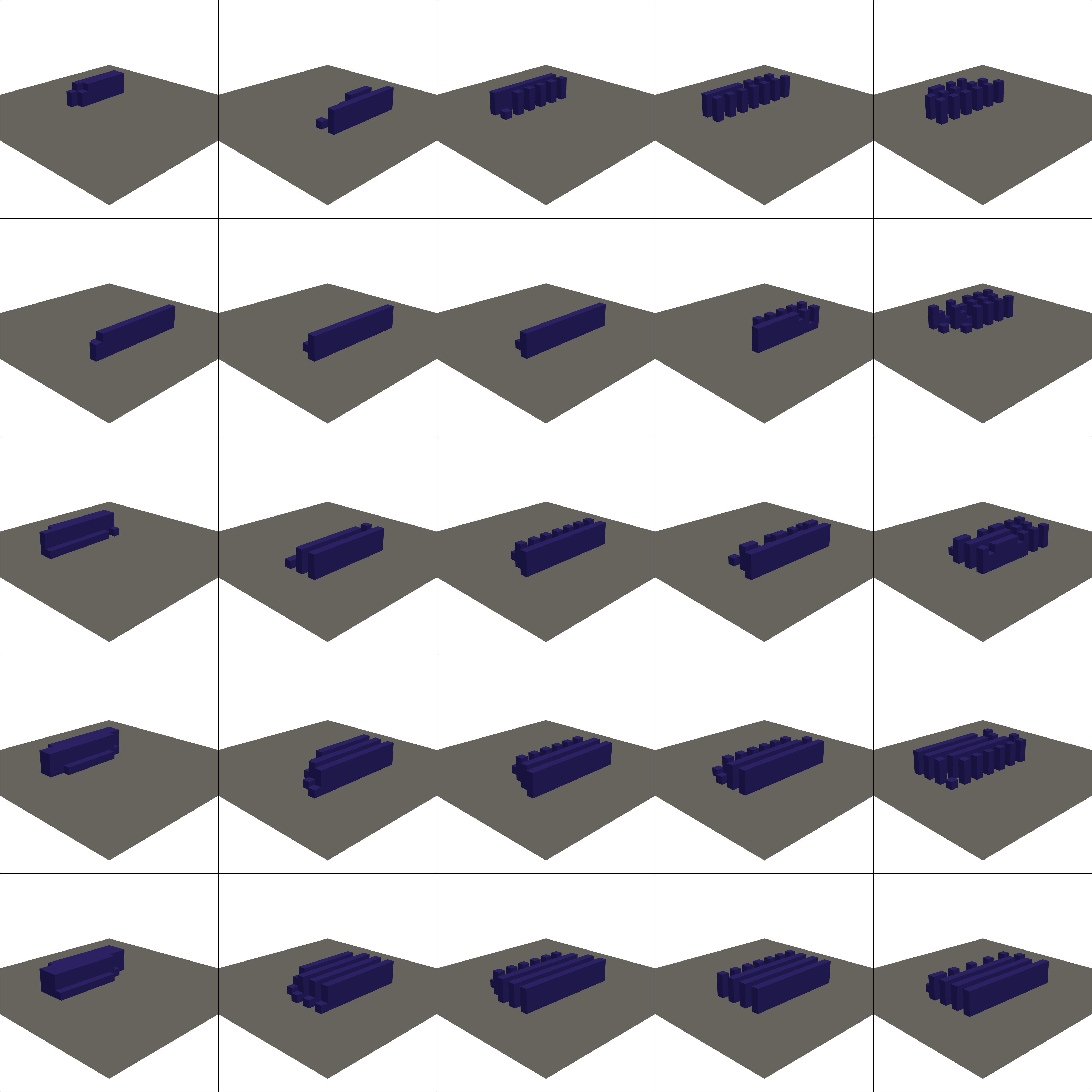}
	\caption{Randomly selected archive with 25 representative solutions using CA encoding.}
	\label{fig:examples:ca}
\end{figure}

\end{document}